\documentclass[letterpaper, 10 pt, conference]{ieeeconf}  

\IEEEoverridecommandlockouts                              

\usepackage{graphicx}   
\usepackage{amsmath}    
\usepackage{amssymb}    
\usepackage{booktabs}   
\usepackage{multirow}   
\usepackage{tabularx}   
\usepackage{threeparttable} 
\usepackage[table]{xcolor}     
\usepackage{soul}       
\usepackage{hyperref}   
\usepackage{cite}       
\usepackage{subcaption}  
\newcommand{\methodname}{DepthCache}

\title{\LARGE \bf
\methodname{}: Depth-Guided Training-Free Visual Token \\
Merging for Vision-Language-Action Model Inference
}

\author{Yuquan Li$^{1}$, Lianjie Ma$^{1}$, Han Ding$^{2}$ and Lijun Zhu$^{1,\dagger}$%
\thanks{$^{\dagger}$Corresponding author.}%
\thanks{$^{1}$School of Artificial Intelligence and Automation,
        Huazhong University of Science and Technology, Wuhan, China.
        {\tt\small \{yuquanli, yingyi1048596, ljzhu\}@hust.edu.cn}}%
\thanks{$^{2}$School of Mechanical Science and Engineering,
        Huazhong University of Science and Technology, Wuhan, China.}%
}

\begin{document}

\maketitle
\thispagestyle{empty}
\pagestyle{empty}

\begingroup
\let\thefootnote\relax\footnotetext{%
This work has been submitted to the IEEE for possible publication.
Copyright may be transferred without notice, after which this version
may no longer be accessible.}
\endgroup


\begin{abstract}
Vision-Language-Action (VLA) models enable generalist robotic manipulation but suffer from high inference latency. This bottleneck stems from the massive number of visual tokens processed by large language backbones. Existing methods either prune or merge tokens uniformly, degrading the spatial reasoning essential for robotic control. We present DepthCache, a training-free framework that leverages depth as a structural prior for visual token compression. It partitions observations into depth-based regions and applies spatially differentiated merge ratios, preserving the near-field workspace while compressing the distant background. To exploit temporal redundancy, DepthCache distributes the merging process across consecutive frames, ensuring consistent representations while reducing per-step computation. A motion-adaptive pipeline further optimizes auxiliary view compression based on end-effector dynamics. The framework requires no model modification, generalizing across diverse VLA architectures. On the LIBERO benchmark, DepthCache achieves up to 1.28$\times$ inference speedup with less than 1\% average success rate degradation across three VLA models ($\pi_{0.5}$, OpenVLA, GR00T), whereas pruning and merging baselines incur 4--24\% degradation at comparable compression. Real-world experiments on a physical manipulator demonstrate that DepthCache enables faster task throughput and more responsive closed-loop control in latency-sensitive scenarios.
\end{abstract}


\section{INTRODUCTION}
\label{sec:introduction}

Vision-Language-Action (VLA) models have emerged as a promising paradigm for generalist robotic manipulation, leveraging the world knowledge encoded in pretrained vision-language backbones to map free-form language instructions and raw visual observations directly to low-level motor commands~\cite{black2024pi0, black2025pi05}.
This integration, however, comes at a steep computational cost: each camera view generates hundreds of visual patch tokens that are subsequently processed by a billion-scale language model, creating inference latencies that fundamentally conflict with the real-time demands of reactive manipulation~\cite{yu2025evlasurvey, black2025rtc}.
As multi-view setups become standard for dexterous tasks, the visual token budget grows proportionally, further widening this latency gap.

\begin{figure}[t]
\centering
\includegraphics[width=\columnwidth]{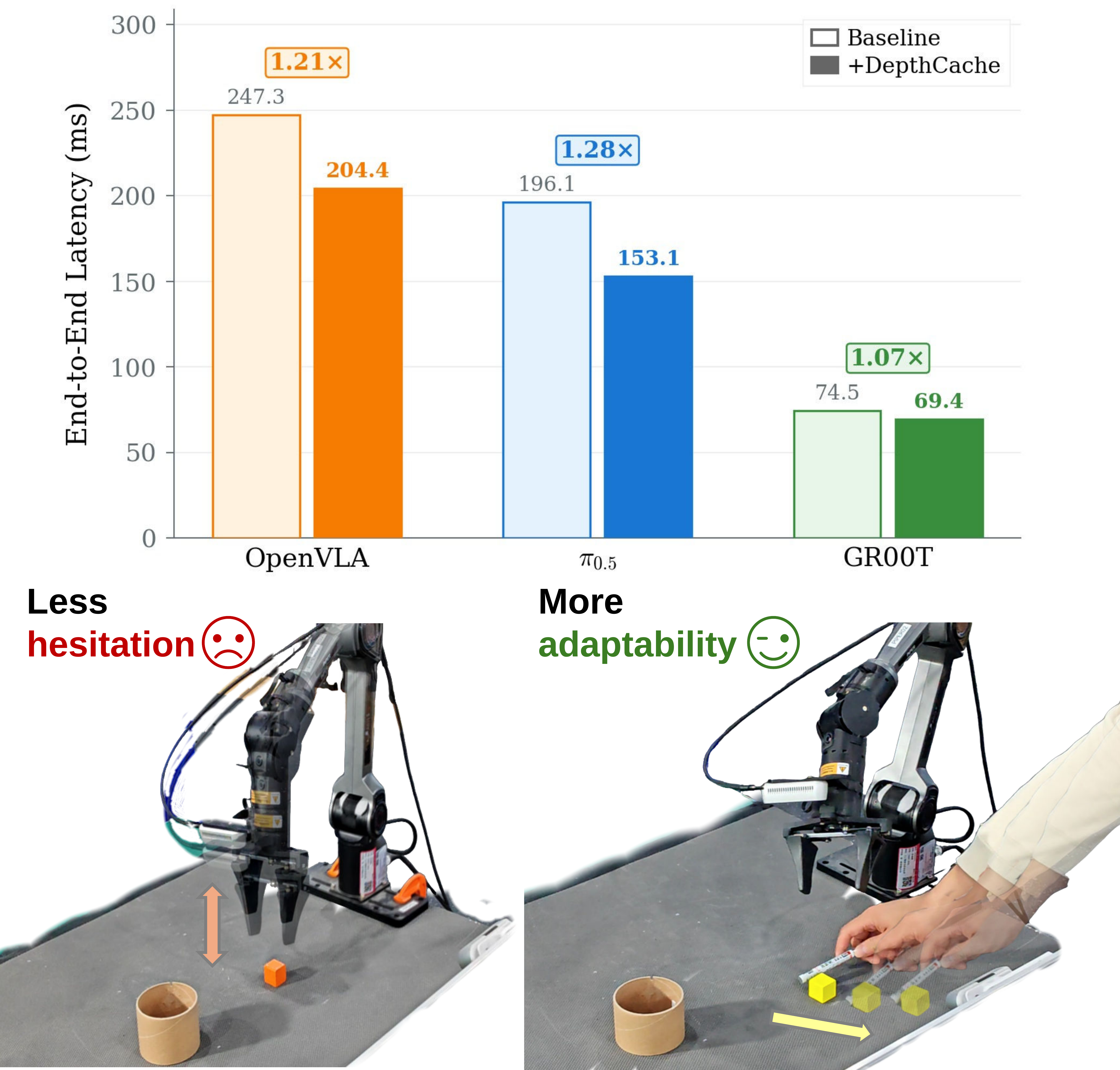}
\caption{\textbf{DepthCache enables training-free inference acceleration for VLA models with negligible performance loss.}
Top: speedup and success rate on LIBERO across three VLA architectures.
Bottom: directly transplanting token merging (ToSA, left) causes hesitation during grasping; DepthCache (right) maintains fluid execution with reduced latency.}
\label{fig:teaser}
\end{figure}

To bridge this gap, a growing body of work has pursued inference-time token \textit{pruning}, which selectively discards visual tokens that score low on task relevance, and reported encouraging speedups~\cite{yang2025efficientvla, xu2025specprune, liu2025adp, sun2025vlacache}.
However, pruning is inherently lossy: regardless of how sophisticatedly tokens are scored, each discarded token severs spatial relationships with its neighbours.
Recent evidence confirms that such destruction degrades fine-grained spatial reasoning in vision-language models~\cite{chen2026nuwa}, a risk amplified in manipulation where precise geometric understanding of the workspace is indispensable.
Token \textit{merging}, which aggregates rather than eliminates, offers an alternative that reduces token count while preserving aggregated information~\cite{bolya2023tome, chen2024tosa}.
Yet existing formulations operate within the vision encoder on a per-layer basis.
Since different VLA models employ distinct vision encoders (e.g., SigLIP, DaViT), each requiring a tailored implementation, such methods lack cross-architecture portability.
Moreover, they apply uniform merge ratios across the entire image, disregarding the spatial disparity between task-relevant near-field objects and distant backgrounds.
An instructive parallel arises from human visuomotor control: during grasping, foveal vision concentrates high-acuity processing on the target and its boundaries, while peripheral regions are perceived at reduced resolution---compressed but never discarded---providing the spatial context essential for online motor coordination~\cite{goodale2006foveal}.
What remains absent is a compression mechanism that, like biological vision, modulates its intensity \textit{both spatially and temporally}: preserving full resolution where precision matters, compressing aggressively where it does not, and maintaining a stable compressed representation across consecutive observations rather than reprocessing every frame from scratch.

A key observation is that robotic manipulation scenes already carry the structural signal needed for such spatially differentiated compression: the \textit{depth map}, available directly from simulators or onboard RGB-D sensors.
Depth naturally delineates the near-field workspace from the distant background and reveals object boundaries---yet no prior work leverages this readily available signal for guiding inference-time compression; existing methods that incorporate depth treat it exclusively as a perceptual input requiring model training~\cite{zuo2025pointvla, qu2025spatialvla, yang2024depthvla}.

Building on this observation, we propose \methodname{}, a training-free visual token compression framework that repurposes depth from a perceptual modality into a \textit{structural prior} for region-aware token merging.
Unprotected tokens are partitioned into depth-based regions, each assigned a merge ratio proportional to its distance from the workspace, compressing distant backgrounds aggressively while preserving near-field detail.
Merging is distributed progressively across the temporal axis, exploiting the high inter-frame visual overlap inherent in sequential robotic control rather than concentrating all reduction within a single forward pass.
A dual protection mechanism, combining the language model's cross-attention as a semantic importance signal with depth-gradient edge detection for geometric boundary preservation, shields task-critical tokens from any compression.
For wrist-mounted cameras, a lightweight motion-adaptive state machine adjusts compression to end-effector dynamics.
The entire framework operates outside the vision encoder, requires no model modification or retraining, and is applicable to any VLA model that accepts visual token inputs.

Our contributions are summarized as follows:
\begin{itemize}
  \item We present \methodname{}, a training-free framework that, to our knowledge, is the first to leverage depth as an external structural prior for spatially differentiated visual token compression in VLA inference.
  \item We introduce a temporally coherent compression pipeline that distributes token merging across consecutive frames and couples depth-proportional region partitioning with dual semantic-geometric protection, eliminating the inter-frame instability and uniform-ratio limitations of existing merge methods.
  \item We validate \methodname{} on three architecturally diverse VLA models ($\pi_{0.5}$, OpenVLA, GR00T) across the LIBERO benchmark, achieving up to 1.28$\times$ inference speedup with less than 1\% average SR degradation. Real-world experiments on a physical manipulator demonstrate 1.33$\times$ speedup across three manipulation tasks and two extended latency-sensitive scenarios.
\end{itemize}


\section{RELATED WORK}
\label{sec:related_work}

\textbf{Visual Token Optimization for VLA Inference.}
The computational bottleneck of VLA models stems largely from the volume of visual tokens processed by the LLM backbone: a single camera view typically produces 256--576 patch tokens, and multi-view setups multiply this proportionally~\cite{yu2025evlasurvey}.
A growing line of work addresses this through inference-time token reduction, predominantly via \textit{pruning}---selectively discarding tokens deemed less relevant to the current task.
These methods generally derive importance scores from one of two sources:
cross-modal attention between language instructions and visual tokens~\cite{yang2025efficientvla, li2025flashvla, zhang2025spvla}, or temporal consistency across consecutive frames~\cite{xu2025specprune, sun2025vlacache}.
Several approaches further condition pruning on the robot's action state, increasing retention during precision-critical phases and pruning more aggressively during coarse motion~\cite{liu2025adp, zhang2025spvla}.
However, pruning irreversibly discards spatial relationships among tokens.
N\"uwa~\cite{chen2026nuwa} quantifies this risk, showing 7--47\% degradation on visual grounding tasks due to disrupted spatial reference frames.
This motivates the exploration of alternatives that reduce token count while \textit{preserving} aggregated information.

\begin{figure*}[t]
\centering
\includegraphics[width=\textwidth]{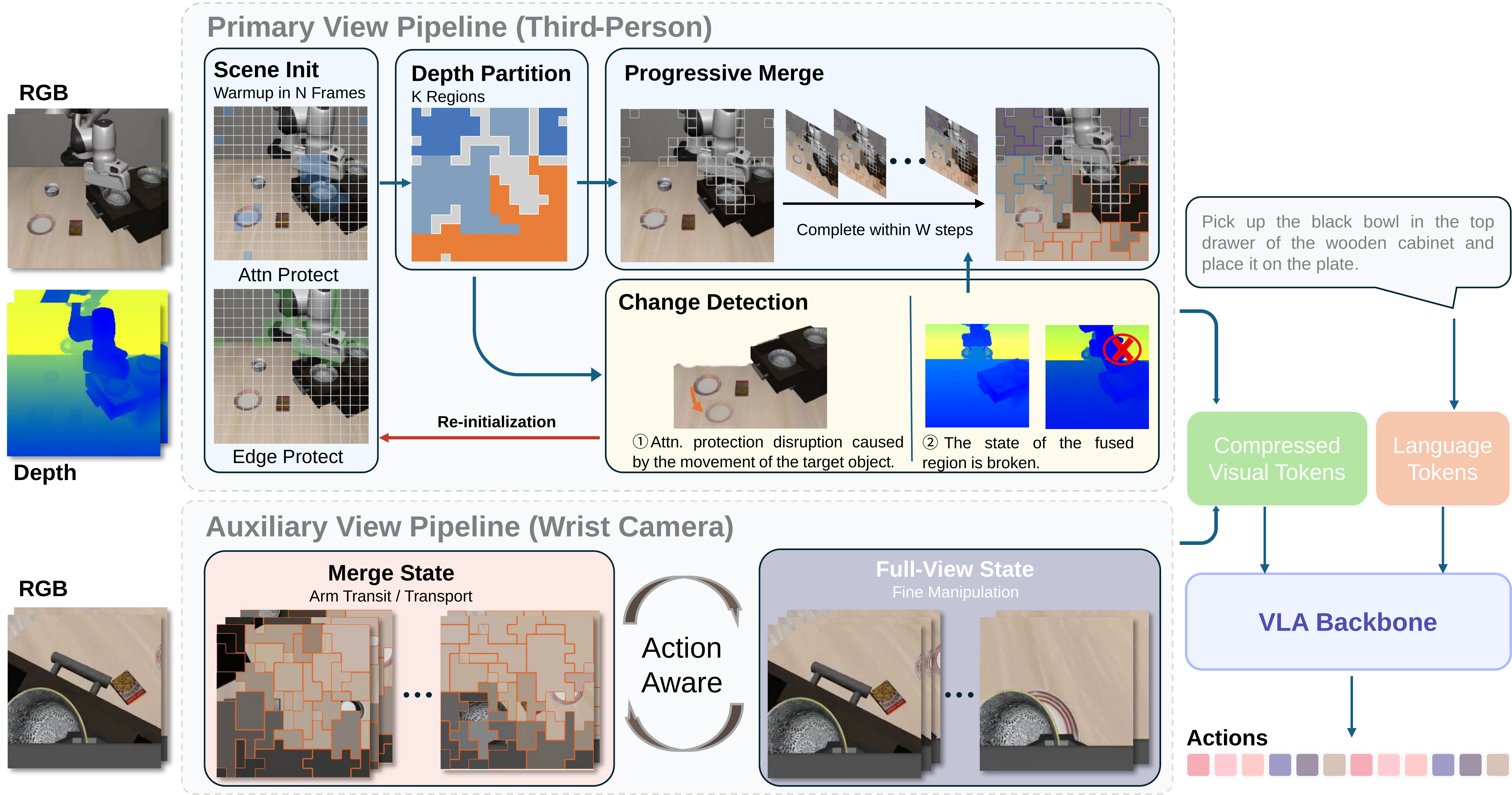}
\caption{\textbf{Overview of \methodname{}.}
The primary view pipeline (top) forms a cyclic process: initialization computes dual protection sets, depth-based partitioning assigns differentiated merge ratios, and progressive merging reduces tokens across frames until scene change triggers re-initialization.
The auxiliary view pipeline (bottom) employs a two-state machine to gate wrist camera compression by end-effector motion.}
\label{fig:overview}
\end{figure*}

\textbf{Depth and 3D Representations in VLA Models.}
A separate but active research direction seeks to enhance VLA models' spatial reasoning by incorporating depth or 3D geometric information.
Existing approaches broadly fall into two paradigms: \textit{explicit 3D input}, where point clouds (e.g., PointVLA~\cite{zuo2025pointvla}) or depth maps~\cite{shridhar20253dcavla, yang2024depthvla, zhu20243dvla} are provided as additional input channels, and \textit{implicit geometric encoding}, where 3D priors are injected through spatial position encoding (e.g., SpatialVLA~\cite{qu2025spatialvla}), stereo feature fusion~\cite{chen2025stereovla}, or knowledge distillation from geometry-aware foundation models~\cite{yuan2025glad, fan2026peafowl}.
Despite their methodological diversity, these approaches share a common assumption: depth serves as a \textit{perceptual} signal---an additional modality that the model must learn to interpret.
Consequently, all require pretraining or fine-tuning to acquire geometric representations, precluding training-free deployment.
No prior work has explored depth as an \textit{external structural prior} for guiding inference-time compression decisions.

\textbf{From Pruning to Merging in VLA Settings.}
The information loss inherent in pruning naturally motivates \textit{token merging}, which reduces token count through aggregation rather than elimination.
ToMe~\cite{bolya2023tome} established the foundational framework via bipartite soft matching within ViT layers, later extended to video~\cite{choi2024vidtldr} and VLA settings~\cite{wang2025teamvla}; ToSA~\cite{chen2024tosa} further incorporated depth-derived spatial tokens to improve merge quality.
However, these methods require architecture-specific modifications to the vision encoder and apply a uniform merge ratio that treats foreground and background identically.
\methodname{} addresses both limitations through depth-guided, spatially differentiated compression that operates entirely outside the vision encoder without retraining (Sec.~\ref{sec:method}).


\section{METHOD}
\label{sec:method}

\subsection{Overview}
\label{sec:overview}

We present \methodname{}, a training-free visual token reduction framework for VLA models.
Our approach is motivated by a key observation: robotic manipulation scenes possess inherent structural regularities that remain largely unexploited by existing token reduction methods~\cite{bolya2023tome}.
Specifically, we identify three such regularities:
\textbf{(i)} \textit{spatial structure}---patches residing at similar depth levels exhibit high inter-token redundancy, while distant regions carry substantially lower information density than the near-field workspace;
\textbf{(ii)} \textit{temporal continuity}---under typical manipulation control frequencies, the majority of scene regions remain visually static across consecutive inference steps;
and \textbf{(iii)} \textit{multi-view asymmetry}---wrist-mounted cameras yield motion-blurred, low-value imagery during arm transit but provide indispensable close-up views during fine manipulation phases.

To systematically exploit these properties, \methodname{} organizes its processing into two camera-specific pipelines (Fig.~\ref{fig:overview}):
\begin{itemize}
  \item \textbf{Primary View Pipeline} (Sec.~\ref{sec:primary}): handles the third-person camera through depth-guided region partitioning, progressive cross-frame token merging, and change-triggered re-initialization.
  \item \textbf{Auxiliary View Pipeline} (Sec.~\ref{sec:auxiliary}): governs wrist cameras via a lightweight state machine that adapts token reduction to end-effector motion dynamics.
\end{itemize}

\subsection{Primary View: Depth-Guided Progressive Token Merging}
\label{sec:primary}

The primary view pipeline constitutes the core of \methodname{}.
It operates as a cyclic process:
\textit{scene initialization} $\rightarrow$ \textit{depth-based partitioning} $\rightarrow$ \textit{progressive merging} $\rightarrow$ \textit{change detection} $\rightarrow$ \textit{re-initialization}, enabling continuous adaptation to evolving task phases.

\subsubsection{Scene Initialization}
\label{sec:initialization}

During the first $N$ warmup frames, the system executes standard forward passes while collecting two complementary protection signals.

\textbf{Attention-based Semantic Protection.}
Since single-frame cross-attention maps can be noisy, we accumulate the LLM's cross-attention weights over image tokens across $N$ warmup frames and average them over all heads to obtain a robust estimate of task-relevant regions.
Patches whose aggregated attention exceeds a dynamic threshold $\tau_{\text{att}} = \mu + \sigma$, where $\mu$ and $\sigma$ are the mean and standard deviation of the per-patch attention scores, form the semantic protection set $\mathcal{P}_{\text{att}}$, identifying task-relevant entities such as the target object and receptacle.

\textbf{Depth-based Geometric Protection.}
We extract geometric boundaries from the depth map via spatial gradient computation.
Patches with depth gradients above threshold $\tau_{\text{edge}}$ form the geometric protection set $\mathcal{P}_{\text{edge}}$, preserving object boundaries and occlusion contours.

\textbf{Combined Protection Set.}
The union $\mathcal{P} = \mathcal{P}_{\text{att}} \cup \mathcal{P}_{\text{edge}}$ integrates semantic and geometric signals: attention captures \textit{what matters} for the task, while depth edges capture \textit{where objects are delineated}.
$\mathcal{P}$ is computed once during initialization and refreshed only upon re-initialization (Sec.~\ref{sec:reinitialization}), avoiding per-frame attention extraction (illustrated in the initialization stage of Fig.~\ref{fig:overview}).

\subsubsection{Depth-Based Region Partitioning}
\label{sec:partitioning}

Following initialization, unprotected patches ($i \notin \mathcal{P}$) are partitioned into $K$ regions via K-Means clustering on their depth values.
Each region $R_k$ is assigned a merge ratio that scales linearly with its mean depth $\bar{d}_k$:
\begin{equation}
  r_k = r_{\min} + (r_{\max} - r_{\min}) \cdot \frac{\bar{d}_k - d_{\min}}{d_{\max} - d_{\min}}
  \label{eq:merge_ratio}
\end{equation}
where $r_{\min}$ and $r_{\max}$ define the merge ratio range.
Protected tokens are assigned $r{=}0$ and remain at full resolution throughout.

\subsubsection{Progressive Token Merging}
\label{sec:progressive_merge}

\methodname{} distributes token merging \textit{across the temporal axis} rather than completing it within a single inference step.

\textbf{Motivation.}
Existing approaches such as ToMe~\cite{bolya2023tome} perform all merging within one forward pass across ViT layers.
While effective for image classification, this one-shot paradigm raises three concerns in sequential manipulation control.
First, merge group assignments are re-computed independently per frame; the resulting inter-frame inconsistencies can cause action hesitation in dynamic control.
Second, concentrating all merges in a single step compounds matching errors within one forward pass, whereas distributing merges across frames allows each step to benefit from fresh visual observations.
Third, at typical control frequencies (5--20\,Hz), consecutive frames share substantial visual overlap that can be exploited for temporally coherent token reduction.

\textbf{Merge Pair Construction.}
Within each depth region $R_k$, we construct merge pairs via bipartite soft matching~\cite{bolya2023tome} on patch embeddings.
Cosine similarity determines greedy assignments between source and target token subsets; the target merge count per region is determined by the depth-assigned ratio $r_k$ and the region size $|R_k|$.
These pairings remain relatively stable across frames to ensure consistent cross-frame merging topology.

\textbf{Temporal Spreading.}
Rather than executing all $m_k$ merges instantaneously, \methodname{} distributes them linearly over a window of $W$ consecutive frames:
\begin{equation}
  m_k^{(t)} = \left\lfloor \frac{\min(t - t_0,\; W)}{W} \cdot m_k \right\rfloor
  \label{eq:temporal_spread}
\end{equation}
where $t_0$ denotes the first post-initialization frame and $m_k$ the target merge count for region $R_k$.
The per-step merge fraction $\eta = 1/W$ controls how quickly compression converges to its target.
Each frame merges only the highest-similarity pairs scheduled for that step, using size-weighted averaging to maintain unbiased token representations across multiple merge rounds.

This design is conceptually analogous to how ToMe distributes merging across ViT layers---but \methodname{} operates \textit{outside} the ViT architecture along the temporal dimension, requiring no model modification and incurring no per-layer synchronization overhead (see the progressive merging stage in Fig.~\ref{fig:overview}).

\textbf{Depth Variation Detection and Dynamic Recovery.}
Progressive merging must account for scene dynamics.
We monitor depth variation for each patch over a sliding window: a patch is deemed static if its depth variation remains below threshold $\epsilon$ across recent frames.
When the fraction of non-static patches within region $R_k$ exceeds threshold $\gamma$, the entire region is restored to full token resolution:
\begin{equation}
  \text{Restore } R_k \;\text{ if }\;
  \frac{|\{i \in R_k \mid \neg\,\text{static}_i\}|}{|R_k|} > \gamma
  \label{eq:restore}
\end{equation}
Restored regions restart progressive merging from scratch upon re-convergence.
This yields adaptive behaviour: static backgrounds are gradually merged, while dynamic regions promptly recover full resolution.

\begin{table*}[!t]
  \centering
  \footnotesize
  \begin{threeparttable}
  \caption{Main results on LIBERO simulation benchmark.}
  \label{tab:main_results}
    \begin{tabularx}{\textwidth}{l *{7}{>{\centering\arraybackslash}X}}
      \toprule
      & \multicolumn{5}{c}{Success Rate (\%)} & \multicolumn{2}{c}{Efficiency} \\
      \cmidrule(lr){2-6} \cmidrule(lr){7-8}
      Method & Spatial & Object & Goal & Long & Avg. & Speedup & $\rho$ \\
      \midrule
      \rowcolor{gray!10}
      OpenVLA                         & 83.3 & 91.2 & 78.6 & 53.7 & 76.7 & 1.00$\times$ & 100 \\
      OpenVLA + FastV                 & 75.8 & 70.3 & 67.3 & 42.5 & 64.0\rlap{\,\scriptsize\color{red!70!black}$_{-12.7}$} & 1.39$\times$ & 75.0 \\
      OpenVLA + SP-VLA$^{\dagger}$    & 80.0 & 82.4 & 73.6 & \textbf{51.6} & 71.9\rlap{\,\scriptsize\color{red!70!black}$_{-4.8}$} & \textbf{1.50$\times$} & dyn.$^{\ddagger}$ \\
      \rowcolor{blue!8}
      OpenVLA + \methodname{}         & \textbf{83.1} & \textbf{90.7} & \textbf{77.8} & 51.2 & \textbf{75.7}\rlap{\,\scriptsize\color{gray}$_{-1.0}$} & 1.21$\times$ & 78.9 \\
      \midrule
      \rowcolor{gray!10}
      $\pi_{0.5}$                     & 99.0 & 97.0 & 99.0 & 96.5 & 97.9 & 1.00$\times$ & 100 \\
      $\pi_{0.5}$ + FastV             & 73.4 & 89.4 & 79.1 & 68.5 & 77.6\rlap{\,\scriptsize\color{red!70!black}$_{-20.3}$} & \textbf{1.30$\times$} & 68.0 \\
      $\pi_{0.5}$ + ToSA              & 79.3 & 82.0 & 71.6 & 62.4 & 73.8\rlap{\,\scriptsize\color{red!70!black}$_{-24.1}$} & 0.94$\times$ & 57.8 \\
      \rowcolor{blue!8}
      $\pi_{0.5}$ + \methodname{}     & \textbf{98.9} & \textbf{97.1} & \textbf{98.3} & \textbf{96.1} & \textbf{97.6}\rlap{\,\scriptsize\color{gray}$_{-0.3}$} & 1.28$\times$ & 68.2 \\
      \midrule
      \rowcolor{gray!10}
      GR00T                           & 91.7 & 97.6 & 93.0 & 90.2 & 93.1 & 1.00$\times$ & 100 \\
      \rowcolor{blue!8}
      GR00T + \methodname{}           & \textbf{93.3} & \textbf{96.4} & \textbf{92.5} & \textbf{89.6} & \textbf{92.9}\rlap{\,\scriptsize\color{gray}$_{-0.2}$} & \textbf{1.07$\times$} & 87.5 \\
      \bottomrule
    \end{tabularx}
    \begin{tablenotes}[flushleft]\footnotesize
      \item All methods are training-free and require no model modification.
        $\rho$: token retention ratio (lower $=$ more reduction).
        $\dagger$: cited from the original publication.
        $\ddagger$: $\rho$ varies dynamically with end-effector velocity.
    \end{tablenotes}
  \end{threeparttable}
\end{table*}

\subsubsection{Re-initialization}
\label{sec:reinitialization}

Scene initialization is not a one-time operation.
Although the target object naturally moves with the end-effector during coordinated manipulation, it may also be displaced independently---e.g., by an external perturbation or after placement at a new location.
To detect such independent displacement, \methodname{} monitors the mean depth change within $\mathcal{P}_{\text{att}}$ while the gripper is \textit{not} actively transporting an object:
\begin{equation}
  \bar{\Delta d} = \frac{1}{|\mathcal{P}_{\text{att}}|}\!\sum_{i \in \mathcal{P}_{\text{att}}}\!\left|d_i^{(t)} - d_i^{(t_{\text{init}})}\right| > \delta_{\text{reinit}}
  \label{eq:reinitialize}
\end{equation}
where $t_{\text{init}}$ denotes the most recent initialization frame.
This condition is evaluated only when the gripper is not carrying an object, ensuring that coordinated arm--object motion does not spuriously trigger re-initialization.
Upon triggering, the system re-enters initialization (Sec.~\ref{sec:initialization}), refreshes $\mathcal{P}$, re-partitions depth regions, and restarts progressive merging.

\subsection{Auxiliary View: Motion-Adaptive Token Reduction}
\label{sec:auxiliary}

The wrist-mounted camera presents fundamentally different dynamics.
During arm transit and object transport, it produces motion-blurred imagery of limited informational value; during fine manipulation, it provides close-up views indispensable for precise control.
\methodname{} employs a two-state machine governed jointly by gripper aperture stability and end-effector motion magnitude, evaluated over the predicted action chunk to enable anticipatory switching.
The system enters \textit{Merge} when the gripper aperture remains stable and end-effector motion is significant, indicating transit or transport phases.
It transitions to \textit{Full-View} when the gripper undergoes an opening or closing transition with low end-effector motion, signaling the onset of a precision-critical grasp or release.
By inspecting the entire action chunk, the system detects upcoming motion trends and switches state preemptively, eliminating temporal lag.

\subsection{Overall Token Budget}
\label{sec:total_budget}

After progressive merging converges, typically within $W$ frames, the system reaches a steady-state token retention ratio.
In our dual-camera configuration, the combined 512 patch tokens are reduced to approximately 300 in steady state, yielding substantial reductions in VLM computation while preserving task-critical visual information.


\section{EXPERIMENTS}
\label{sec:experiments}

We evaluate \methodname{} across multiple VLA architectures on both simulated and real-world manipulation tasks.
Our experiments are designed to answer the following research questions:
\begin{itemize}
  \item \textbf{RQ1}: Can \methodname{} reduce VLA inference latency while preserving task success rates across different model architectures?
  \item \textbf{RQ2}: Does \methodname{} transfer effectively to real-world robotic manipulation?
  \item \textbf{RQ3}: What is the contribution of each design component to the overall performance?
\end{itemize}

\subsection{Experimental Setup}
\label{sec:exp_setup}

\textbf{Simulation Benchmark.}
We evaluate on LIBERO~\cite{liu2024libero}, a widely adopted manipulation benchmark comprising four task suites (Spatial, Object, Goal, Long) with 40 tasks of increasing difficulty.
Each task is evaluated over 100 episodes; we report the average success rate (SR) per suite and the grand average.

\textbf{VLA Baselines.}
To validate the generality of \methodname{} as a model-agnostic compression framework, we apply it to three architecturally diverse VLA models:
(1)~$\pi_{0.5}$~\cite{black2025pi05} (3.3B), pairing a SigLIP vision encoder with a Gemma language backbone and a flow-matching action expert;
(2)~OpenVLA~\cite{kim2024openvla} (7B), fusing SigLIP and DINOv2 encoders with a Llama~2 backbone for autoregressive action prediction;
and (3)~GR00T~\cite{bjorck2025groot} (2.2B), adopting an Eagle vision-language model with a Diffusion Transformer action head.
These models differ in vision encoders, language backbones, and action generation paradigms, providing a rigorous testbed for cross-architecture generality.
All checkpoints are sourced from Hugging Face and fine-tuned on LIBERO.

\textbf{Metrics.}
We report:
(i)~task success rate (SR, \%),
(ii)~end-to-end inference latency per action step (ms),
(iii)~inference speedup relative to the uncompressed baseline ($\times$),
and (iv)~average token compression ratio ($\rho$, fraction of tokens retained after compression).
Latency is measured as the wall-clock time from image observation to action output, averaged over all evaluation episodes.

\textbf{Real-World Platform.}
For real-world validation, we deploy on a PIPER 6-DoF manipulator with two Intel RealSense D435 RGB-D cameras providing an external static view and a wrist-mounted eye-in-hand view.
Depth observations are acquired directly from the onboard stereo depth sensor.

\textbf{Implementation Details.}
Both simulation and real-world experiments are conducted on a single NVIDIA RTX 4090 GPU (24~GB).
Unless otherwise noted, \methodname{} uses $N{=}5$ warmup frames, $K{=}3$ depth clusters, and a progressive merge window of $W{=}5$ steps.

\subsection{Simulation Results}
\label{sec:sim_results}

To contextualize \methodname{}'s design, we compare against baselines spanning pruning and merging paradigms (Table~\ref{tab:main_results}).
For pruning, we evaluate FastV~\cite{chen2024fastv} on both $\pi_{0.5}$ and OpenVLA, and include SP-VLA~\cite{zhang2025spvla} results on OpenVLA.
FastV drops average SR by 20.3\% on $\pi_{0.5}$ and 12.7\% on OpenVLA, with particularly severe losses on spatial tasks (Spatial $-25.6\%$ on $\pi_{0.5}$; Object $-20.9\%$ on OpenVLA).
SP-VLA, which adds action-aware scheduling to semantic pruning, achieves a higher speedup ($1.50\times$) but still incurs a $4.8\%$ average SR loss.
These results confirm that pruning visual tokens, regardless of the selection criterion, systematically degrades manipulation performance by destroying spatial context.
We note that ADP~\cite{liu2025adp} augments pruning with action-aware gating on OpenVLA-OFT, a parallel-decoding variant with a different baseline, precluding direct numerical comparison; its pruning core is the same mechanism evaluated via FastV.
For merging, ToSA~\cite{chen2024tosa} applies uniform intra-ViT merging on $\pi_{0.5}$, reducing SR by 24.1\% with no speedup ($0.94\times$) due to integration overhead.
Three structural limitations explain this result: a uniform merge ratio across all patches, per-frame re-partitioning that introduces temporal discontinuity, and frequent merging of the gripper with the target object during grasping.

\textbf{Cross-architecture results.}
Across all three architectures, \methodname{} achieves $1.07$--$1.28\times$ speedup with less than $1\%$ average SR loss.
In every case, \methodname{} loses less than $1\%$ average SR while all pruning and merging baselines exceed $4\%$ degradation at comparable or lower $\rho$.

\subsection{Real-World Experiments}
\label{sec:real_world}

To validate that \methodname{} transfers to physical manipulation, we evaluate on five real-world tasks using a PIPER robotic arm with dual RealSense D435 cameras (Sec.~\ref{sec:exp_setup}).
We organize the evaluation into two tiers: three core tasks that assess fundamental manipulation competence under compression, and two extended scenarios that probe the practical benefits of reduced inference latency.

\subsubsection{Core Tasks}
\label{sec:core_tasks}

\textbf{Task Design.}
We select three tasks that exercise distinct aspects of \methodname{}:
\begin{enumerate}
  \item \textbf{Pick \& Place}: grasp a cube from the table and place it at a designated bin.
    Tests basic manipulation robustness under token compression.
  \item \textbf{Stack Blocks}: pick up a cube and stack it on top of another cube.
    Requires precise spatial alignment, exercising the dual protection mechanism that preserves object boundaries during fine positioning.
  \item \textbf{Open Drawer \& Place}: open a drawer, grasp an object, and place it inside.
    A multi-phase task whose scene-structure changes (drawer opening, object relocation) exercise the depth variation detection mechanism, testing whether token reduction adapts correctly across distinct manipulation phases.
\end{enumerate}
Each task is evaluated over 20 trials for both the baseline and \methodname{}-augmented $\pi_{0.5}$ policy.

\begin{table}[!t]
  \centering
  \footnotesize
  \caption{Real-world core task results with $\pi_{0.5}$.}
  \label{tab:real_world}
    \begin{tabular*}{\columnwidth}{@{\extracolsep{\fill}}l c c}
      \toprule
      \multirow{2}{*}{Task} & \multicolumn{2}{c}{Successes / 20} \\
      \cmidrule(lr){2-3}
           & Baseline & + Ours \\
      \midrule
      Pick \& Place        & 20 / 20 & 20 / 20 \\
      Stack Blocks         & 18 / 20 & 17 / 20 \\
      Drawer \& Place      & 17 / 20 & 15 / 20 \\
      \midrule
      Total                & 55 / 60 & 52 / 60 \\
      \midrule
      Avg.\ Latency (ms)  & 191 & 143 \\
      Speedup              & 1.00$\times$ & 1.33$\times$ \\
      \bottomrule
    \end{tabular*}
\end{table}

\begin{figure}[t]
  \centering
  \begin{subfigure}[b]{0.48\columnwidth}
    \includegraphics[width=\textwidth]{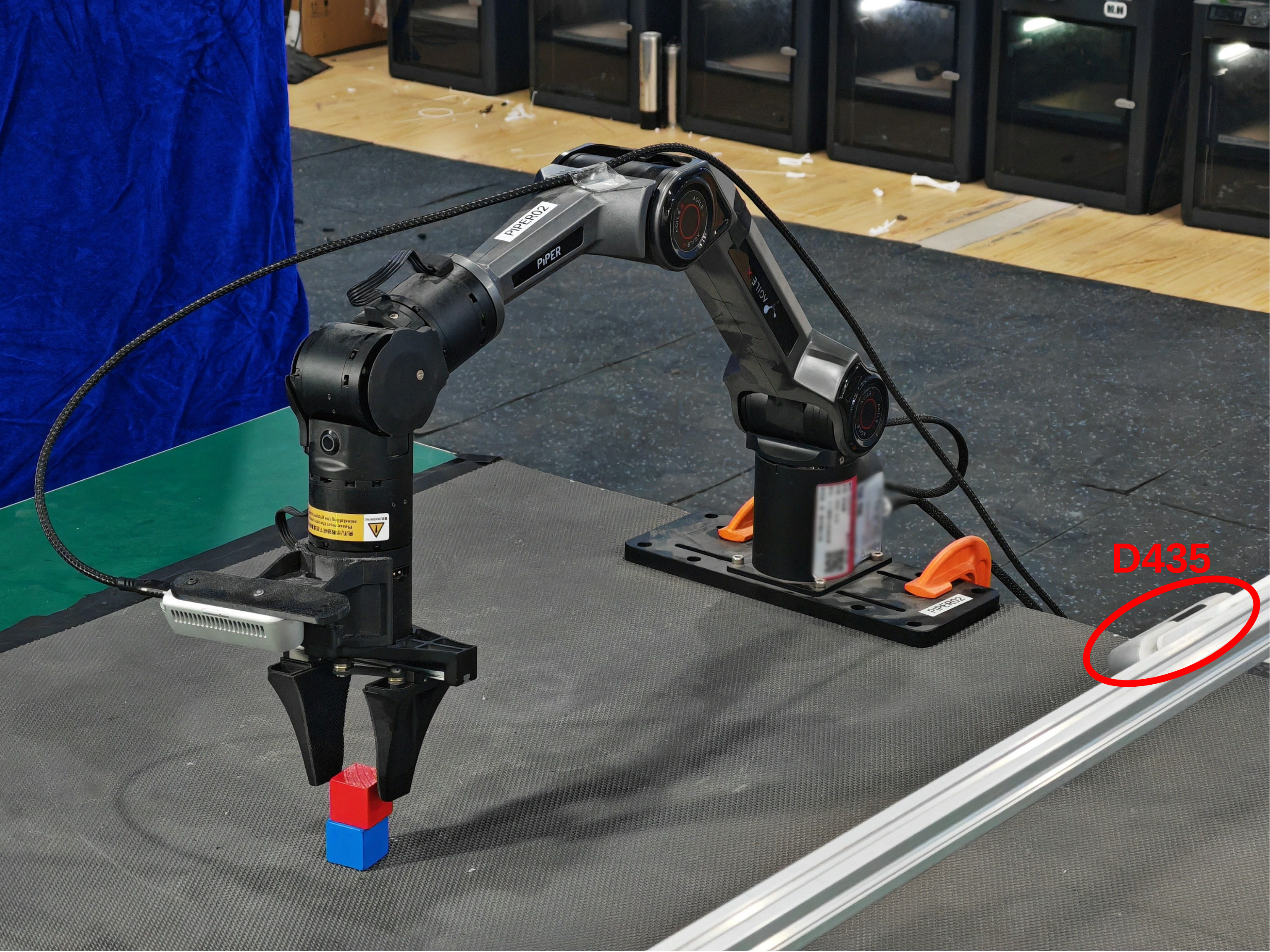}
    \caption{Stack Blocks}
  \end{subfigure}\hfill
  \begin{subfigure}[b]{0.48\columnwidth}
    \includegraphics[width=\textwidth]{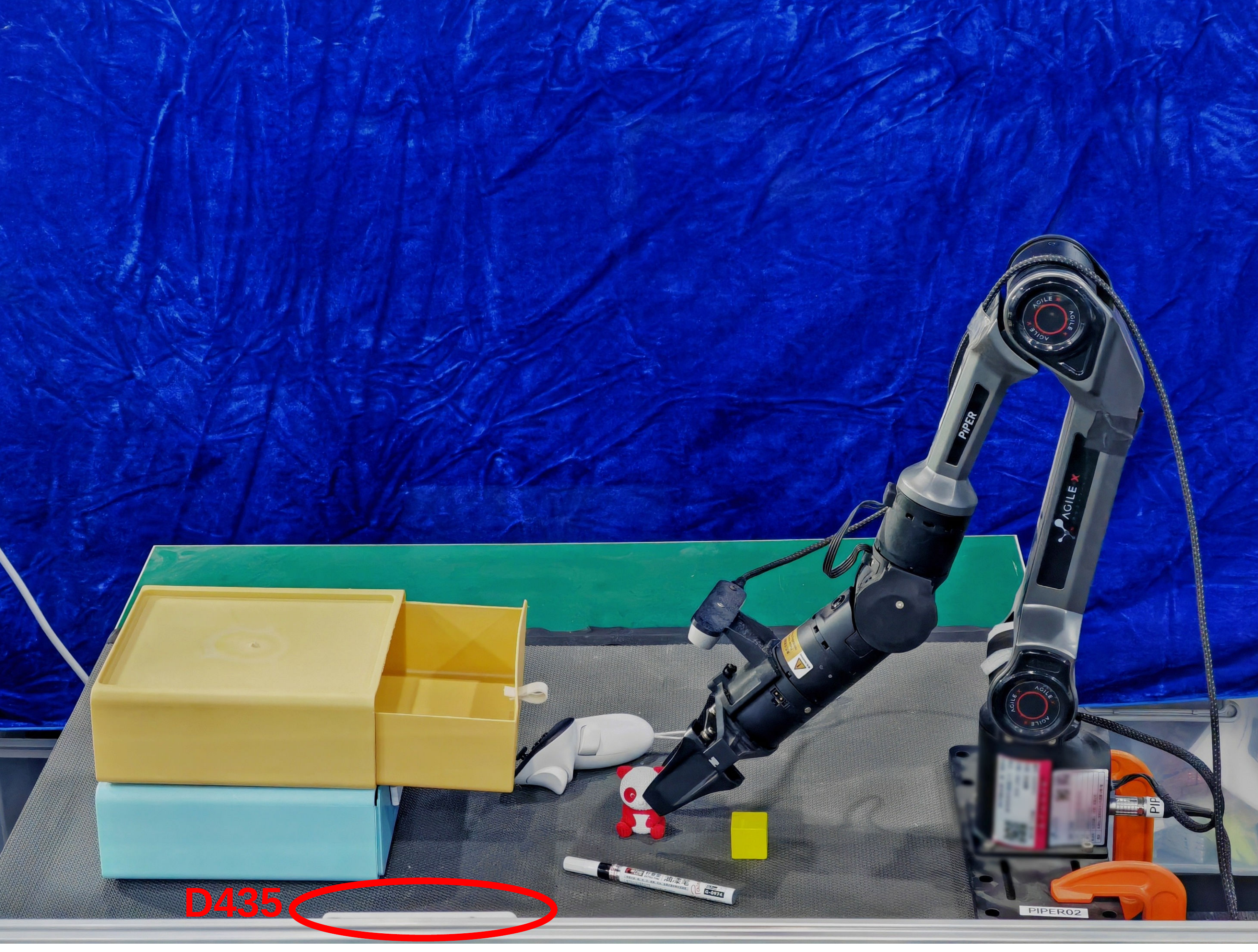}
    \caption{Open Drawer \& Place}
  \end{subfigure}
  \caption{\textbf{Real-world core task execution} with \methodname{}-augmented $\pi_{0.5}$. Perturbation recovery is shown in Fig.~\ref{fig:teaser}.}
  \label{fig:real_world_tasks}
\end{figure}

As shown in Table~\ref{tab:real_world}, \methodname{} maintains comparable success rates to the uncompressed baseline across all three core tasks (Fig.~\ref{fig:real_world_tasks}), with an average inference speedup of 1.33$\times$.

\subsubsection{Extended Scenarios}
\label{sec:extended_scenarios}

\begin{figure*}[!t]
  \centering
  \includegraphics[width=\textwidth]{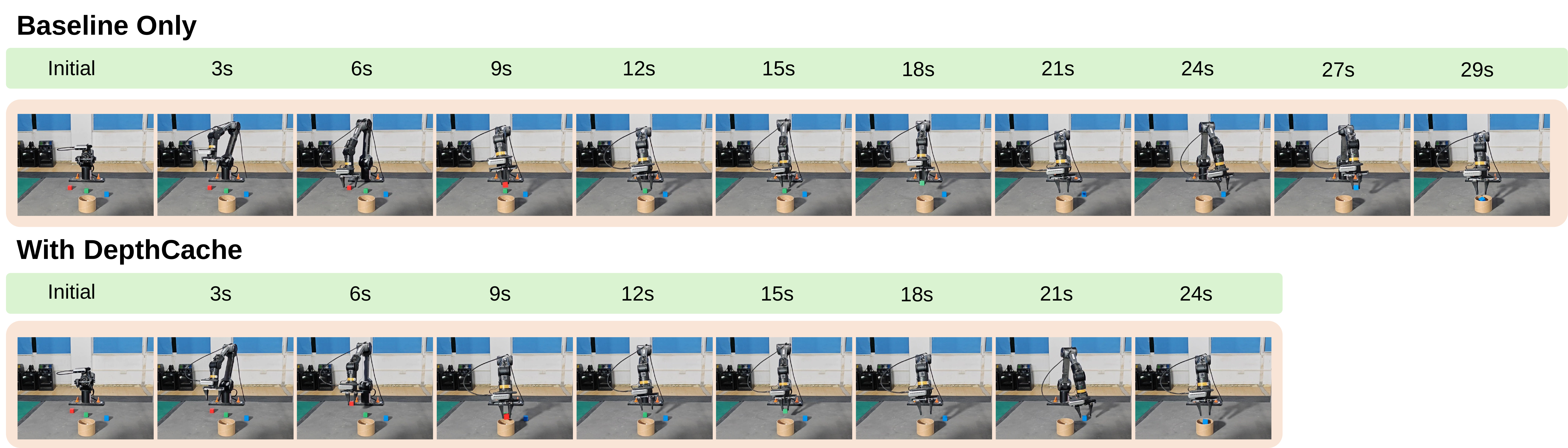}
  \caption{\textbf{Sequential multi-object sorting comparison.} 
    Top: baseline $\pi_{0.5}$. Bottom: $\pi_{0.5}$ + \methodname{}.
    The time gap accumulates across successive pick-and-place cycles.}
  \label{fig:multi_object}
\end{figure*}

\textbf{Sequential Multi-Object Sorting.}
We place 3 colored cubes on the table and instruct the robot to sort them into designated bin one by one.
This task chains multiple pick-and-place cycles, accumulating latency differences into a visible gap in total task completion time (Fig.~\ref{fig:multi_object}).
Each trial is recorded as a side-by-side synchronized video; we report the average total elapsed time across 15 trials.

\textbf{Perturbation Recovery.}
During a pick-and-place trial, while the arm is moving toward the target cube, we manually push the cube approximately 2--3\,cm away from its original position.
The reduced inference latency provided by \methodname{} enables more frequent re-planning, allowing the policy to detect and correct for the displacement more promptly.

\begin{table}[!b]
  \centering
  \footnotesize
  \caption{Extended scenario results (15 trials each).}
  \label{tab:extended}
    \begin{tabular*}{\columnwidth}{@{\extracolsep{\fill}}l c c c c}
      \toprule
      \multirow{2}{*}{Scenario} & \multicolumn{2}{c}{Successes / 15} & \multicolumn{2}{c}{Time (s)} \\
      \cmidrule(lr){2-3} \cmidrule(lr){4-5}
               & Baseline & + Ours & Baseline & + Ours \\
      \midrule
      Multi-Object Sorting   & 15 / 15 & 13 / 15 & 28.6 & 22.1\\
      Perturbation Recovery  & 11 / 15 & 12 / 15 & 17.4 & 13.7 \\
      \bottomrule
    \end{tabular*}
\end{table}

As shown in Table~\ref{tab:extended}, \methodname{} reduces multi-object sorting time by 22.7\% (28.6\,s $\rightarrow$ 22.1\,s) while maintaining a comparable success rate, confirming that per-step inference savings accumulate into a substantial throughput gain over repeated manipulation cycles.
In the perturbation scenario, the faster re-planning loop improves recovery time by 21.3\% (17.4\,s $\rightarrow$ 13.7\,s) and slightly raises the success rate (11/15 $\rightarrow$ 12/15), indicating that the policy benefits from receiving corrective observations sooner.

\subsection{Ablation Study}
\label{sec:ablation}

We conduct ablation experiments on LIBERO using $\pi_{0.5}$ to quantify the contribution of each component.

\subsubsection{Component Analysis}
\label{sec:component_ablation}


Table~\ref{tab:ablation_components} shows that the two \emph{spatial} mechanisms, Depth Partitioning and Progressive Merge, have the largest impact on task performance ($\Delta = -18.2\%$ and $-16.6\%$), while the remaining components serve complementary roles in safety, adaptation, and multi-view acceleration.
Across all ablations, Object and Long suites are consistently the most affected, as both require sustained precision over fine-grained spatial features and are therefore most sensitive to token information loss.

Depth Partitioning is the single most critical component.
Replacing depth-proportional allocation with a uniform merge ratio removes the inductive bias that near-field tokens carry denser task-relevant information than distant background.
Without this prior, compression erodes gripper-object geometry and target features equally, causing severe failure on Object ($-22.0\%$) and Long ($-25.0\%$).
Progressive Merge ranks second: one-shot application of the full merge ratio introduces a step discontinuity in the visual conditioning signal, which is especially destructive for Object tasks ($-29.0\%$) where grasp trajectories depend on smooth inter-chunk feature consistency.

Among the remaining components, Dual Protection removal yields a moderate SR drop ($-7.3\%$) while \emph{increasing} speedup to $1.48\times$, confirming that protection trades compression headroom for safety.
Re-initialization primarily benefits Spatial ($-7.1\%$) and Long ($-6.7\%$) tasks where scene geometry evolves during execution, with negligible effect on static-scene Object tasks.
The Auxiliary View Pipeline contributes a speedup gain from $1.06\times$ to $1.28\times$ with near-zero SR impact ($+0.2\%$), validating the auxiliary-view state machine as a safe acceleration mechanism.

\subsubsection{Parameter Sensitivity}
\label{sec:param_sensitivity}

\begin{figure}[!b]
  \centering
  \includegraphics[width=\columnwidth]{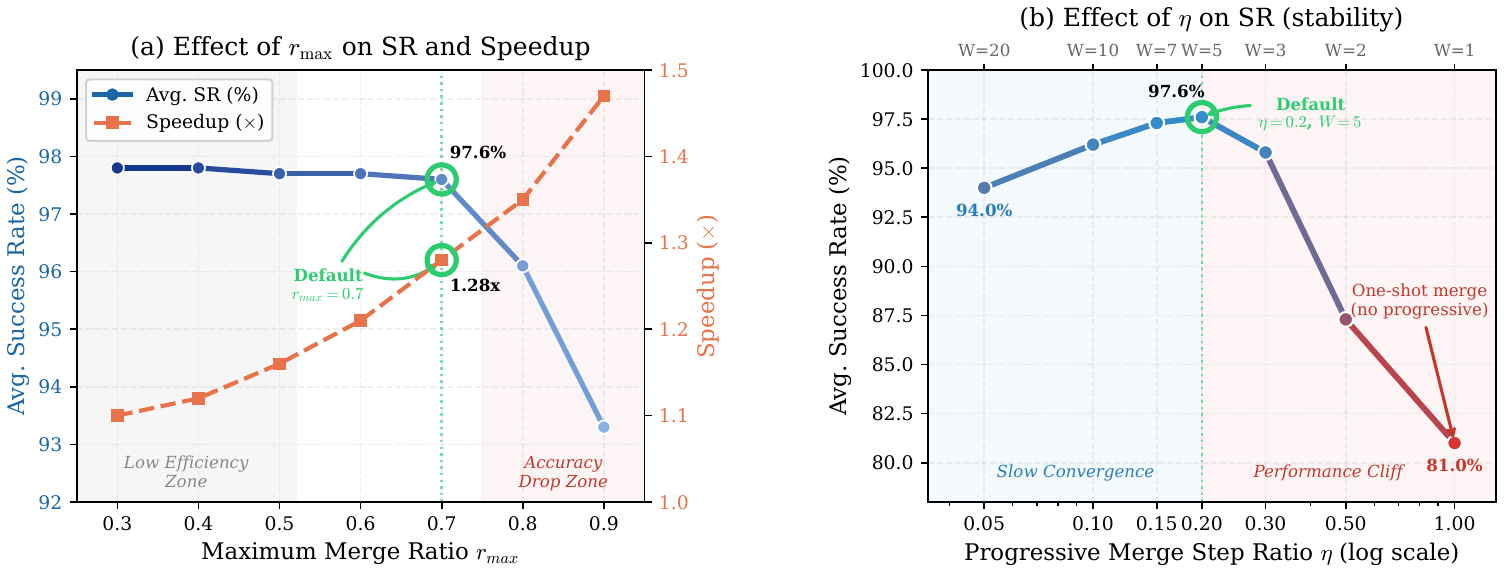}
  \caption{\textbf{Parameter sensitivity analysis} on $\pi_{0.5}$ (LIBERO). (a) Effect of maximum merge ratio $r_{\max}$ on success rate and speedup. (b) Effect of progressive merge step ratio $\eta$ on success rate (log-scale $x$-axis). Default values ($r_{\max}=0.7$, $\eta=0.2$) are highlighted.}
  \label{fig:param_sensitivity}
\end{figure}

We investigate the sensitivity of \methodname{} to its two key hyperparameters (Fig.~\ref{fig:param_sensitivity}).

\textbf{Maximum merge ratio} $r_{\max}$ modulates the overall compression rate via the depth-proportional mapping $r_k \in [r_{\min}, r_{\max}]$.
Across a broad range ($r_{\max} \in [0.3, 0.7]$), SR remains nearly invariant, demonstrating a wide and flat safe operating plateau.
This robustness stems from the dual protection mechanism, which unconditionally shields task-critical tokens from merging, and the depth-proportional assignment, which concentrates compression on low-value distant regions.
Degradation appears only beyond $r_{\max} \approx 0.7$, where even the moderately compressed near-field regions begin to lose spatial resolution.
The default $r_{\max} = 0.7$ thus represents the Pareto-optimal trade-off: maximum speedup at the boundary of the safe zone.

\textbf{Progressive merge step ratio} $\eta$ controls the transition speed to target compression ($W \approx 1/\eta$ frames).
In VLA inference, each action chunk is conditioned on visual tokens from the preceding step;
a step change in token count between chunks introduces a discontinuity in the conditioning signal, disrupting the smoothness of generated trajectories.
Progressive merging amortizes this transition, allowing consecutive action chunks to be conditioned on gradually varying inputs.
At the extreme ($\eta = 1.0$, one-shot merge), the system degenerates to the \textit{w/o Progressive Merge} ablation (SR $81.0\%$), confirming that temporal smoothing is essential rather than optional.

\begin{table*}[!t]
  \centering
  \footnotesize
  \begin{threeparttable}
  \caption{Ablation study on \methodname{} components using $\pi_{0.5}$.}
  \label{tab:ablation_components}
    \begin{tabularx}{\textwidth}{l *{8}{>{\centering\arraybackslash}X}}
      \toprule
      \multirow{2}{*}{Configuration} & \multicolumn{6}{c}{Success Rate (\%)} & \multicolumn{2}{c}{Efficiency} \\
      \cmidrule(lr){2-7} \cmidrule(lr){8-9}
      & Spatial & Object & Goal & Long & Avg. & $\Delta$ & Speedup & $\rho$ \\
      \midrule
      Full \methodname{} & 98.9 & 97.1 & 98.3 & 96.1 & 97.6 & --- & 1.28$\times$ & 68.2 \\
      \midrule
      \quad w/o Depth Partitioning & 92.9 & 75.1 & 78.4 & 71.1 & 79.4 & -18.2 & 1.34$\times$ & 67.5 \\
      \quad w/o Dual Protection & 96.0 & 85.7 & 92.1 & 87.2 & 90.3 & -7.3 & 1.48$\times$ & 55.6 \\
      \quad w/o Progressive Merge & 94.7 & 68.1 & 82.5 & 78.6 & 81.0 & -16.6 & 1.32$\times$ & 65.4 \\
      \quad w/o Re-initialization & 91.8 & 97.2 & 92.6 & 89.4 & 92.8 & -4.8 & 1.25$\times$ & 72.5 \\
      \quad w/o Auxiliary View Pipeline & 98.2 & 98.0 & 98.6 & 96.2 & 97.8 & +0.2 & 1.06$\times$ & 88.0 \\
      \bottomrule
    \end{tabularx}
    \begin{tablenotes}[flushleft]\footnotesize
      \item Each row removes one component from the full method.
        $\Delta$: change in average SR relative to the full method (percentage points).
    \end{tablenotes}
  \end{threeparttable}
\end{table*}


\section{CONCLUSION}
\label{sec:conclusion}

We showed that depth, commonly treated as a perceptual input for VLA models, can instead serve as a structural prior for inference-time visual token compression.
The resulting framework, \methodname{}, applies spatially differentiated and temporally coherent merging without model modification, generalizing across three architecturally distinct VLA backbones with less than 1\% success rate degradation on the LIBERO benchmark.
Real-world experiments further confirm that inference savings translate into faster task throughput and more responsive closed-loop control on a physical manipulator.
Two principal limitations remain.
First, as a purely inference-time method, \methodname{} compresses visual tokens but does not accelerate the action decoding stage (e.g., flow-matching denoising), bounding achievable speedup by Amdahl's law.
Second, our evaluation covers three VLA architectures on the LIBERO benchmark and a single 6-DoF arm; extending to more diverse benchmarks, embodiments, and real-world settings would further validate generality.
Incorporating depth-aware compression into the training pipeline and combining token merging with complementary strategies such as KV-cache optimization are promising directions for lifting these ceilings.


\bibliographystyle{IEEEtran}
\bibliography{references}

\begin{thebibliography}{10}
\providecommand{\url}[1]{#1}
\csname url@samestyle\endcsname
\providecommand{\newblock}{\relax}
\providecommand{\bibinfo}[2]{#2}
\providecommand{\BIBentrySTDinterwordspacing}{\spaceskip=0pt\relax}
\providecommand{\BIBentryALTinterwordstretchfactor}{4}
\providecommand{\BIBentryALTinterwordspacing}{\spaceskip=\fontdimen2\font plus
\BIBentryALTinterwordstretchfactor\fontdimen3\font minus
  \fontdimen4\font\relax}
\providecommand{\BIBforeignlanguage}[2]{{%
\expandafter\ifx\csname l@#1\endcsname\relax
\typeout{** WARNING: IEEEtran.bst: No hyphenation pattern has been}%
\typeout{** loaded for the language `#1'. Using the pattern for}%
\typeout{** the default language instead.}%
\else
\language=\csname l@#1\endcsname
\fi
#2}}
\providecommand{\BIBdecl}{\relax}
\BIBdecl

\bibitem{black2024pi0}
K.~Black, N.~Brown, D.~Driess, A.~Esmail, M.~Equi, C.~Finn, N.~Fusai, L.~Groom,
  K.~Hausman, B.~Ichter \emph{et~al.}, ``{$\pi_0$}: A vision-language-action
  flow model for general robot control,'' \emph{arXiv preprint
  arXiv:2410.24164}, 2024.

\bibitem{black2025pi05}
K.~Black, N.~Brown, J.~Darpinian, K.~Dhabalia, D.~Driess, A.~Esmail, M.~Equi,
  C.~Finn, N.~Fusai, M.~Y. Galliker \emph{et~al.}, ``{$\pi_{0.5}$}: A
  vision-language-action model with open-world generalization,'' in
  \emph{Annual Conference on Robot Learning (CoRL)}, 2025.

\bibitem{yu2025evlasurvey}
Z.~Yu, B.~Wang, P.~Zeng, H.~Zhang, J.~Zhang, Z.~Wang, L.~Gao, J.~Song, N.~Sebe,
  and H.~T. Shen, ``A survey on efficient vision-language-action models,''
  \emph{arXiv preprint arXiv:2510.24795}, 2025.

\bibitem{black2025rtc}
K.~Black, M.~Y. Galliker, and S.~Levine, ``Real-time execution of action
  chunking flow policies,'' in \emph{Advances in Neural Information Processing
  Systems (NeurIPS)}, 2025.

\bibitem{yang2025efficientvla}
Y.~Yang, Y.~Wang, Z.~Wen, L.~Zhong, C.~Zou, Z.~Zhang, C.~Wen, and L.~Zhang,
  ``{EfficientVLA}: Training-free acceleration and compression for
  vision-language-action models,'' in \emph{Advances in Neural Information
  Processing Systems (NeurIPS)}, 2025.

\bibitem{xu2025specprune}
H.~Wang, J.~Xu, Y.~Xiang, J.~Pan, Y.~Zhou, Y.-L. Li, and G.~Dai,
  ``{SpecPrune-VLA}: Accelerating vision-language-action models via
  action-aware self-speculative pruning,'' \emph{arXiv preprint
  arXiv:2509.05614}, 2025.

\bibitem{liu2025adp}
X.~Pei, Y.~Chen, S.~Xu, Y.~Wang, Y.~Shi, and C.~Xu, ``Action-aware dynamic
  pruning for efficient vision-language-action manipulation,'' in
  \emph{International Conference on Learning Representations (ICLR)}, 2026.

\bibitem{sun2025vlacache}
S.~Xu, Y.~Wang, C.~Xia, D.~Zhu, T.~Huang, and C.~Xu, ``{VLA-Cache}: Efficient
  vision-language-action manipulation via adaptive token caching,'' in
  \emph{Advances in Neural Information Processing Systems (NeurIPS)}, 2025.

\bibitem{chen2026nuwa}
Y.~Huang, F.~Ma, Y.~Shao, J.~Guo, Z.~Yu, L.~Cui, and Q.~Tian, ``N{\"u}wa:
  Mending the spatial integrity torn by {VLM} token pruning,'' in
  \emph{International Conference on Learning Representations (ICLR)}, 2026.

\bibitem{bolya2023tome}
D.~Bolya, C.-Y. Fu, X.~Dai, P.~Zhang, C.~Feichtenhofer, and J.~Hoffman, ``Token
  merging: Your {ViT} but faster,'' in \emph{International Conference on
  Learning Representations (ICLR)}, 2023.

\bibitem{chen2024tosa}
H.-W. Huang, W.~Chai, K.-M. Chen, C.-Y. Yang, and J.-N. Hwang, ``{ToSA}: Token
  merging with spatial awareness,'' in \emph{IEEE/RSJ International Conference
  on Intelligent Robots and Systems (IROS)}, 2025, pp. 9654--9660.

\bibitem{goodale2006foveal}
M.~A. Goodale and A.~D. Milner, ``Action without perception in human vision,''
  \emph{Cognitive Neuropsychology}, vol.~25, no. 7-8, pp. 891--919, 2008.

\bibitem{zuo2025pointvla}
C.~Li, J.~Wen, Y.~Peng, Y.~Peng, F.~Feng, and Y.~Zhu, ``{PointVLA}: Injecting
  the {3D} world into vision-language-action models,'' \emph{arXiv preprint
  arXiv:2503.07511}, 2025.

\bibitem{qu2025spatialvla}
D.~Qu, H.~Song, Q.~Chen, Y.~Yao, X.~Ye, Y.~Ding, Z.~Wang, J.~Gu, B.~Zhao,
  D.~Wang \emph{et~al.}, ``{SpatialVLA}: Exploring spatial representations for
  visual-language-action model,'' \emph{arXiv preprint arXiv:2501.15830}, 2025.

\bibitem{yang2024depthvla}
T.~Yuan, Y.~Liu, C.~Lu, Z.~Chen, T.~Jiang, and H.~Zhao, ``Depthvla: Enhancing
  vision-language-action models with depth-aware spatial reasoning,''
  \emph{arXiv preprint arXiv:2510.13375}, 2025.

\bibitem{li2025flashvla}
X.~Tan, Y.~Yang, P.~Ye, J.~Zheng, B.~Bai, X.~Wang, J.~Hao, and T.~Chen, ``Think
  twice, act once: Token-aware compression and action reuse for efficient
  inference in vision-language-action models,'' \emph{arXiv preprint
  arXiv:2505.21200}, 2025.

\bibitem{zhang2025spvla}
Y.~Li, Y.~Meng, Z.~Sun, K.~Ji, C.~Tang, J.~Fan, X.~Ma, S.-T. Xia, Z.~Wang, and
  W.~Zhu, ``{SP-VLA}: A joint model scheduling and token pruning approach for
  {VLA} model acceleration,'' in \emph{International Conference on Learning
  Representations (ICLR)}, 2026.

\bibitem{shridhar20253dcavla}
V.~Bhat, Y.-H. Lan, P.~Krishnamurthy, R.~Karri, and F.~Khorrami, ``{3D CAVLA}:
  Leveraging depth and {3D} context to generalize vision language action models
  for unseen tasks,'' \emph{arXiv preprint arXiv:2505.05800}, 2025.

\bibitem{zhu20243dvla}
H.~Zhen, X.~Qiu, P.~Chen, J.~Yang, X.~Yan, Y.~Du, Y.~Hong, and C.~Gan,
  ``3{D}-{VLA}: A 3{D} vision-language-action generative world model,'' in
  \emph{Proc. International Conference on Machine Learning (ICML)}, 2024, pp.
  61\,229--61\,245.

\bibitem{chen2025stereovla}
S.~Deng, M.~Yan, Y.~Zheng, J.~Su, W.~Zhang, X.~Zhao, H.~Cui, Z.~Zhang, and
  H.~Wang, ``Stereovla: Enhancing vision-language-action models with stereo
  vision,'' \emph{arXiv preprint arXiv:2512.21970}, 2025.

\bibitem{yuan2025glad}
M.~Guo, M.~Cao, J.~Tao, R.~Xu, Y.~Yan, X.~Liang, I.~Laptev, and X.~Chang,
  ``Glad: Geometric latent distillation for vision-language-action models,''
  \emph{arXiv preprint arXiv:2512.09619}, 2025.

\bibitem{fan2026peafowl}
Q.~Fan, Z.~Li, Y.~Lu, W.~Chen, Q.~Shen, X.-x. Long, Y.~Cai, T.~Lu, S.~Wang, and
  X.~Cao, ``{PEAfowl}: Perception-enhanced multi-view vision-language-action
  for bimanual manipulation,'' \emph{arXiv preprint arXiv:2601.17885}, 2026.

\bibitem{choi2024vidtldr}
J.~Choi, S.~Lee, J.~Chu, M.~Choi, and H.~J. Kim, ``vid-tldr: Training free
  token merging for light-weight video transformer,'' in \emph{Proc. IEEE/CVF
  Conf. Computer Vision and Pattern Recognition (CVPR)}, 2024, pp.
  18\,771--18\,781.

\bibitem{wang2025teamvla}
Y.~Ye, J.~Ma, J.~Cen, and Z.~Lu, ``Token expand-merge: Training-free token
  compression for vision-language-action models,'' \emph{arXiv preprint
  arXiv:2512.09927}, 2025.

\bibitem{liu2024libero}
B.~Liu, Y.~Zhu, C.~Gao, Y.~Feng, Q.~Liu, Y.~Zhu, and P.~Stone, ``{LIBERO}:
  Benchmarking knowledge transfer for lifelong robot learning,'' in
  \emph{Advances in Neural Information Processing Systems (NeurIPS)}, vol.~36,
  2023, pp. 44\,776--44\,791.

\bibitem{kim2024openvla}
M.~J. Kim, K.~Pertsch, S.~Karamcheti, T.~Xiao, A.~Balakrishna, S.~Nair,
  R.~Rafailov, E.~P. Foster, P.~R. Sanketi, Q.~Vuong, T.~Kollar, B.~Burchfiel,
  R.~Tedrake, D.~Sadigh, S.~Levine, P.~Liang, and C.~Finn, ``{OpenVLA}: An
  open-source vision-language-action model,'' in \emph{Proc. Conference on
  Robot Learning (CoRL)}, 2025, pp. 2679--2713.

\bibitem{bjorck2025groot}
J.~Bjorck, F.~Casta{\~n}eda, N.~Cherniadev, X.~Da, R.~Ding, L.~Fan, Y.~Fang,
  D.~Fox, F.~Hu, S.~Huang \emph{et~al.}, ``{GR00T N1}: An open foundation model
  for generalist humanoid robots,'' \emph{arXiv preprint arXiv:2503.14734},
  2025.

\bibitem{chen2024fastv}
L.~Chen, H.~Zhao, T.~Cai, Y.~Cao, Y.~Chen, and Y.~Lu, ``An image is worth 1/2
  tokens after layer 2: Plug-and-play inference acceleration for large
  vision-language models,'' in \emph{European Conference on Computer Vision
  (ECCV)}, 2024.

\end{thebibliography}

\end{document}